\documentclass[letterpaper, 10 pt, conference]{ieeeconf}  
\IEEEoverridecommandlockouts                              
\usepackage{amsmath} % assumes amsmath package installed
\usepackage{amssymb} % assumes amsmath package installed

\usepackage{float}
\usepackage{graphicx}
\usepackage{amsmath}

\newcommand{\vect}[1]{\boldsymbol{#1}}
\newcommand{\mat}[1]{\boldsymbol{#1}}

\newcommand{\diffs}[3]{\frac{\partial^2 #1}{
\ifx#2#3 
\partial #2^2
\else
\partial #2 \partial #3
\fi
}}
\newcommand{\av}{\vect{a}}

\newcommand{\hv}{\vect{h}}

\newcommand{\pv}{\vect{p}}

\newcommand{\qv}{{\vect{q}}}

\newcommand{\rv}{{\vect{r}}}

\newcommand{\uv}{\vect{u}}
\newcommand{\vv}{\vect{v}}

\newcommand{\wv}{\vect{w}}

\newcommand{\xv}{\vect{x}}

\newcommand{\muv}{\vect{\mu}}

\newcommand{\phiv}{\vect{\phi}}

\newcommand{\tauv}{\vect{\tau}}

\newcommand{\Hm}{\mat{H}}
\newcommand{\Jm}{\mat{J}}

\newcommand{\Rm}{\mat{R}}

\newcommand{\Tm}{\mat{T}}

\usepackage{hyperref}
\usepackage[export]{adjustbox}
\usepackage{xcolor}
\usepackage[normalem]{ulem}
\usepackage{booktabs}
\let\labelindent\relax
\usepackage{enumitem}
\usepackage[symbol]{footmisc}

\usepackage{balance} 
\usepackage{duckuments}
\def\IntroductionPhoto{\centering \includegraphics[width=0.45\textwidth]{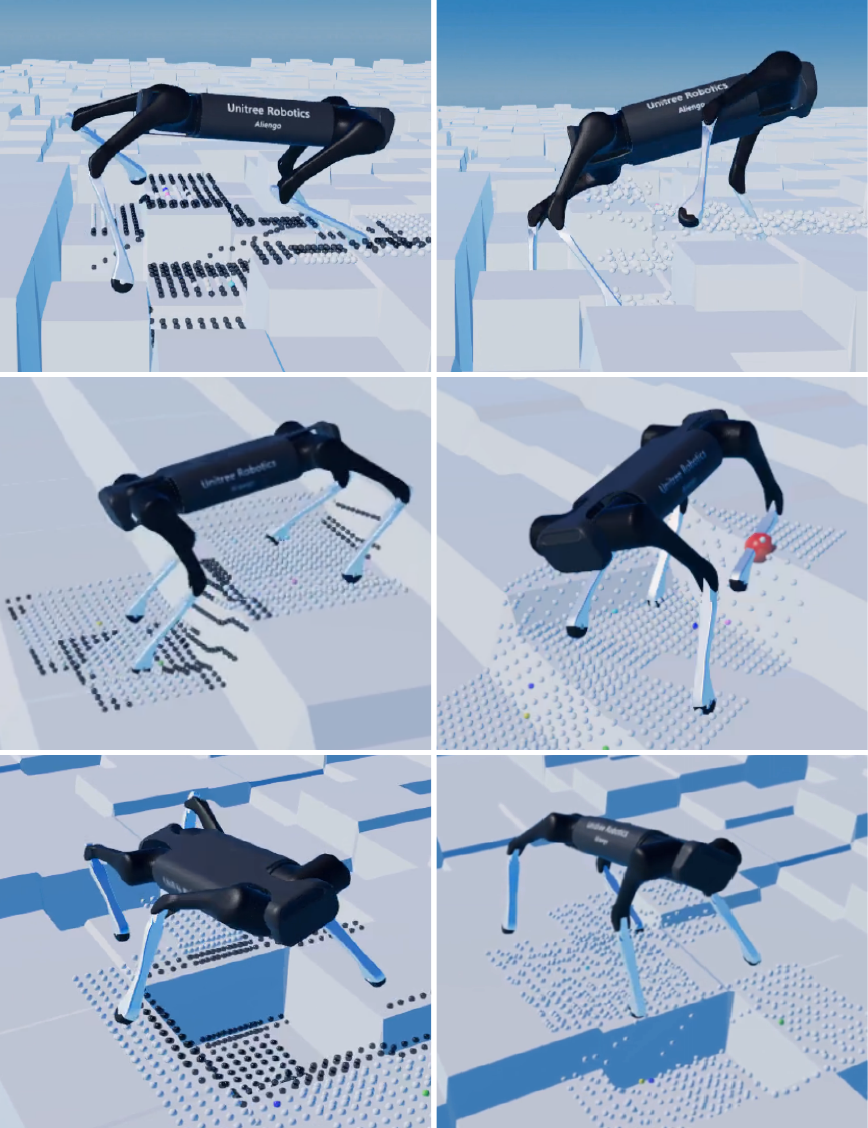}
}

\def\BlockDiagram{
\centerline{\includegraphics[width=0.975\textwidth]{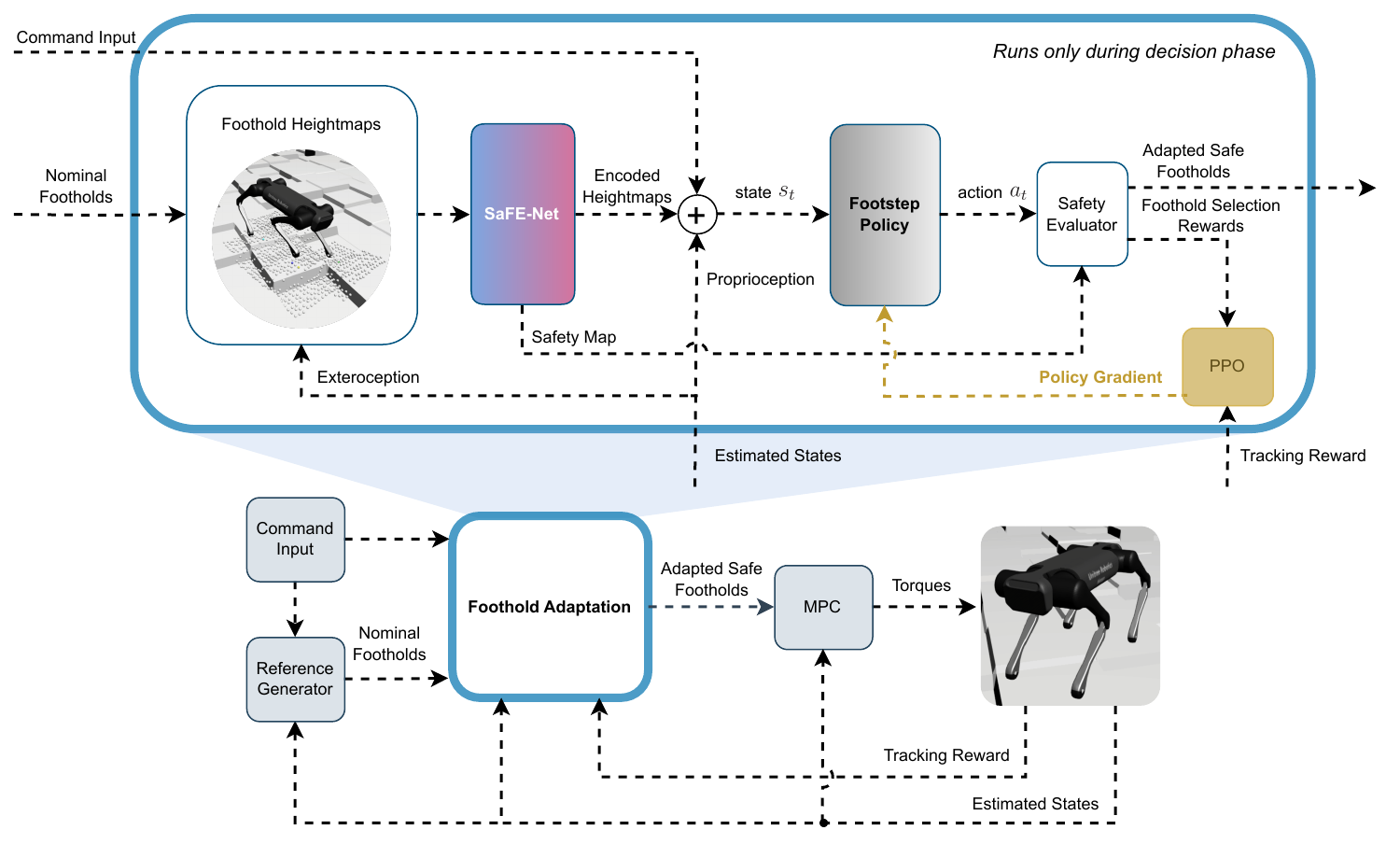}}
}

\def\SafeNet{
\centerline{\includegraphics[width=0.95\textwidth]{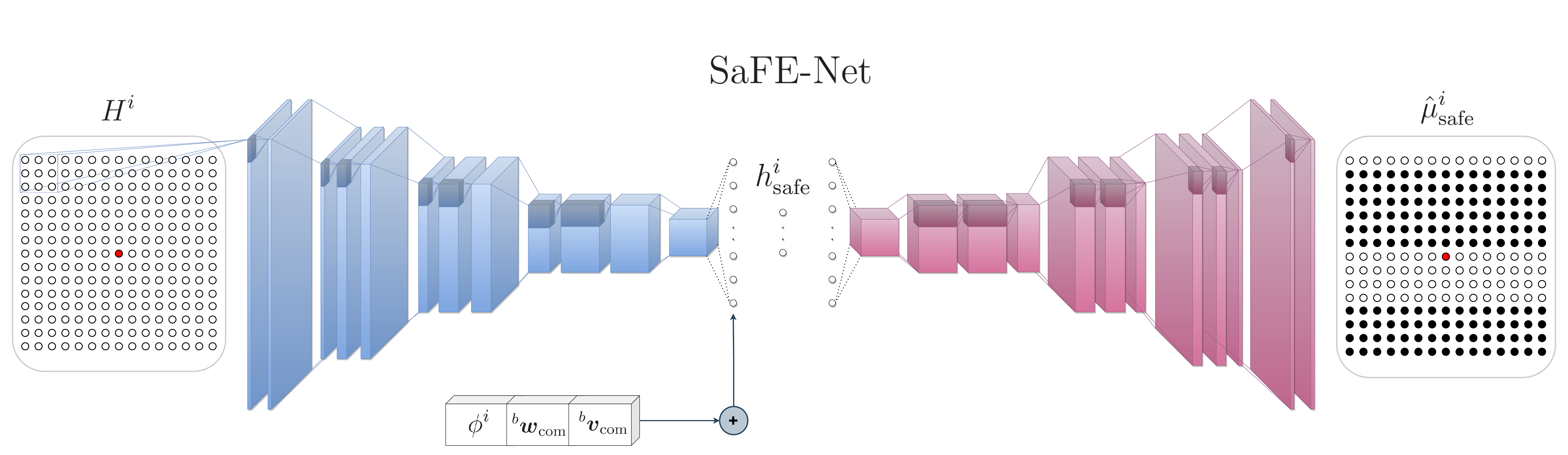}}
}

\def\TrackingCost{
\includegraphics[width=1\textwidth]{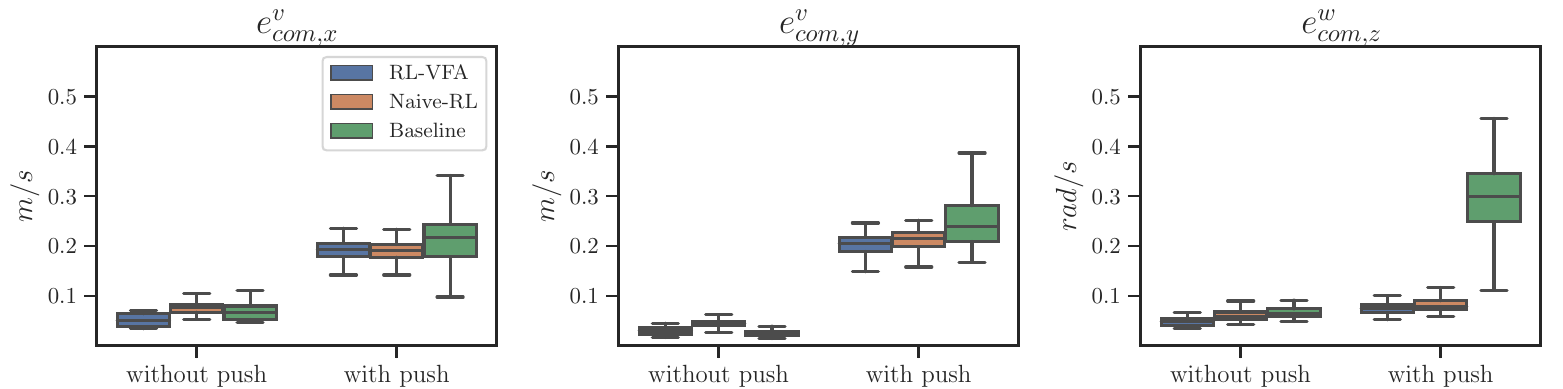}
}

\def\DisturbanceEval{
\centerline{\includegraphics[width=0.5\textwidth]{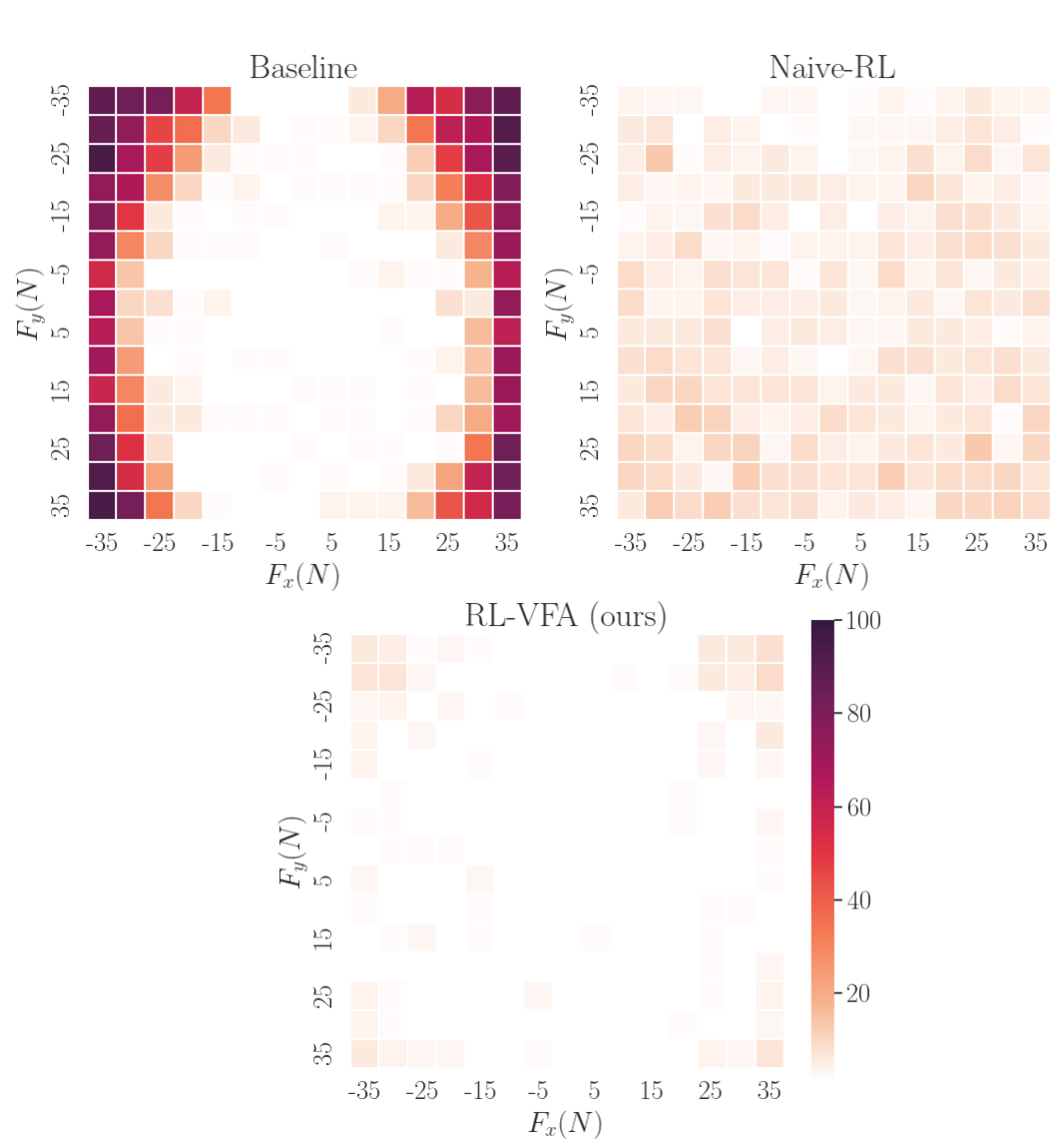}}
}

\def\SuccessRate{
\centerline{\includegraphics[width=0.5\textwidth]{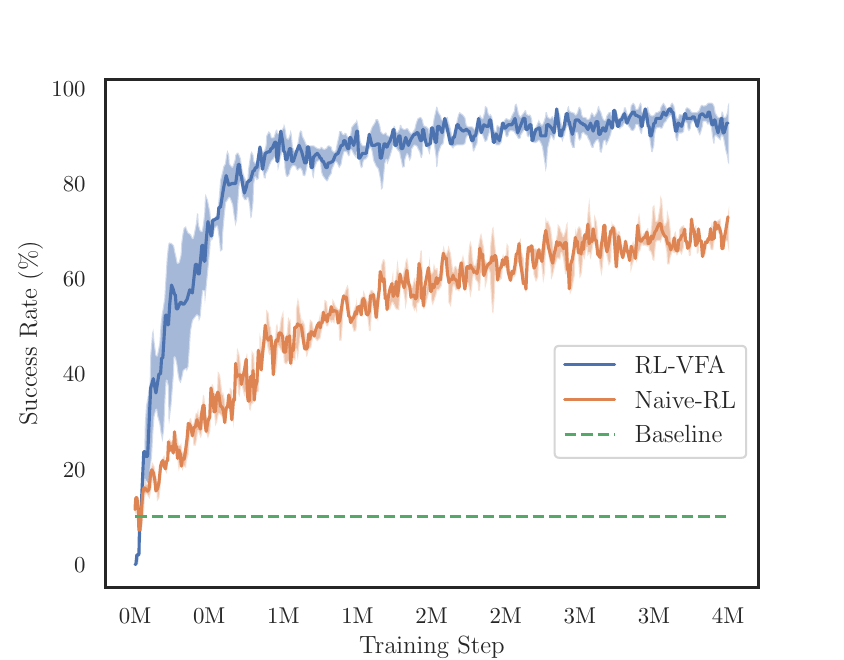}}
}

\usepackage{xcolor}
\usepackage{eso-pic}
\AddToShipoutPictureBG*{%
  \AtPageUpperLeft{%
    \setlength{\unitlength}{1mm}%
    \put(-9.5,-9){\makebox(\paperwidth,0)[c]{\parbox{0.9\textwidth}{2023 IEEE-RAS 22nd International Conference on Humanoid Robots (Humanoids), pp. 1-8}}}
  }
}

\AddToShipoutPictureBG*{%
  \AtPageUpperLeft{%
    \setlength{\unitlength}{1mm}%
    \put(-9.5,-14){\makebox(\paperwidth,0)[c]{\parbox{0.9\textwidth}{DOI: \href{https://ieeexplore.ieee.org/document/10375218}{10.1109/Humanoids57100.2023.10375218}}}}
  }
}
    
\title{ \LARGE \bf
SafeSteps: Learning Safer Footstep Planning Policies \\ for Legged Robots via Model-Based Priors}

\author{Shafeef Omar, Lorenzo Amatucci, Victor Barasuol,  Giulio Turrisi, Claudio Semini%
\thanks{The authors are with the Dynamic Legged Systems Laboratory, Istituto Italiano di Tecnologia (IIT), Genova, Italy.
E-mail:\{name.lastname\}@iit.it.}%
}%

\begin{document}

\maketitle
\thispagestyle{empty}
\pagestyle{empty}

\begin{abstract}
We present a footstep planning policy for quadrupedal locomotion that is able to directly take into consideration a-priori safety information in its decisions. At its core, a learning process analyzes terrain patches, classifying each landing location by its kinematic feasibility, shin collision, and terrain roughness. This information is then encoded into a small vector representation and passed as an additional state to the footstep planning policy, which furthermore proposes only safe footstep location by applying a masked variant of the Proximal Policy Optimization algorithm. The performance of the proposed approach is shown by comparative simulations and experiments on an electric quadruped robot walking in different rough terrain scenarios. We show that violations of the above safety conditions are greatly reduced both during training and the successive deployment of the policy, resulting in an inherently safer footstep planner. Furthermore, we show how, as a byproduct, fewer reward terms are needed to shape the behavior of the policy, which in return is able to achieve both better final performances and sample efficiency. 
%The videos of the obtained results can be found at \href{https://sites.google.com/view/safe-steps-rl}{https://sites.google.com/view/safe-steps-rl}.
\\

\end{abstract}

\section{Introduction}
\label{sec:introduction}
Quadruped robots are increasingly entering new application domains, such as inspection, construction, and rescue. To be valuable in such real-world scenarios, robots should be able to map the surroundings via their onboard sensors, such as cameras or lidar, in order to choose the best footstep landing location autonomously and avoid stepping on potentially unsafe regions (e.g., collapsing surfaces, stairs edges, etc.) which can inevitably bring the systems to failure. Along with visual information, an additional component needed to traverse such harsh scenarios is the ability to perform a reactive stepping strategy for recovery, e.g., modifying the robot's footstep at need in the presence of an external disturbance. This requirement further increases the complexity of the problem, given the need to explicitly consider the robot's dynamics along with vision. In both the case of planning and recovery, the resulting chosen footsteps need to be kinematically feasible to be actuated by the robot, and the resulting motion needs to be collision-free in order to avoid additional ground reaction forces that can destabilize the motion. 

\begin{figure}[!t]
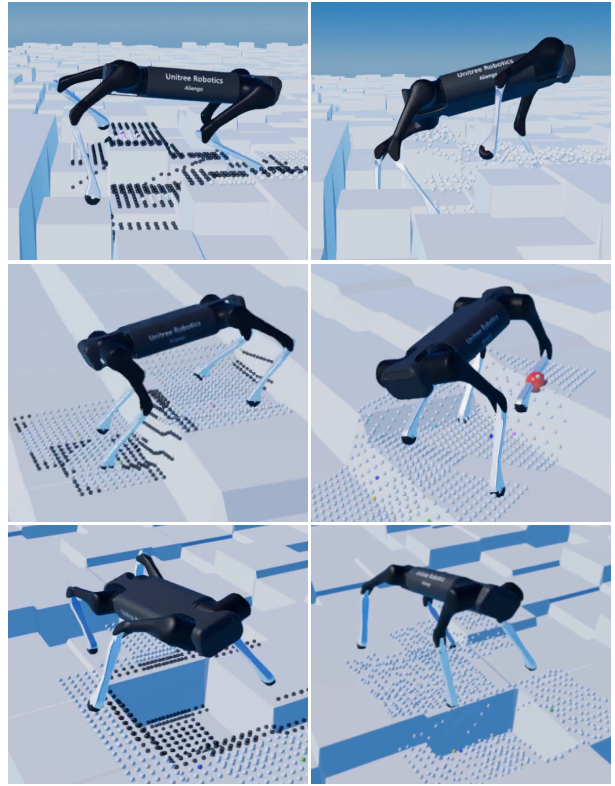

\vspace{5pt}
\IntroductionPhoto
\caption{Snapshots of Aliengo~\cite{aliengo} walking over irregular terrains with our method (left) where footholds that bring safety violations are prohibited at the action level (black dots), and with a naive RL policy (right) where these violations are only discouraged during training. Starting from the top, we depict a kinematic violation, a shin collision, and a slippage due to the foot being placed over a terrain edge, by the naive RL policy.}
\label{fig:scenarios}
\vspace{-18pt}
\end{figure}

Given the difficulty of achieving all these conditions concurrently, footstep placement techniques are still heavily studied in the robotic community. In~\cite{vision1}, the authors presented one of the first applications of terrain awareness locomotion to enhance the traversing capabilities of
a quadruped robot.  A set of heuristics was analyzed to select the best foothold locations, discarding possible candidates near edges, slopes, or holes.
This idea was then extended in~\cite{vision2} by considering body collisions and introducing a supervised learning regression technique to ease the classification problem. Still, given the needed computational time, the method was unsuitable for reactive and agile motion. This reactiveness was then achieved by performing regression via fast Convolutional Neural Networks in~\cite{villarreal19ral}. In this case, candidate footsteps were chosen by analyzing terrain features, the kinematic limits of the robot, and possible shin collision. Given the achieved computational speed, the robot was able to withstand various disturbances applied during its motion. This method, which we refer to as Visual Foothold Adaptation (VFA), will be used as the core component of this work.

More recently, optimization-based control techniques, such as Model Predictive Control~\cite{book_mpc} (MPC), were coupled with vision-based footstep correction~\cite{mpc_vision1},~\cite{mpc_vision2},~\cite{mpc_safenet},~\cite{mpc_vision3},~\cite{bratta}  to obtain highly-dynamic and optimal motions. Noticeably, in~\cite{mpc_vision2} and~\cite{mpc_safenet}, the footstep position was considered as an additional control variable to be further optimized by the MPC, taking into consideration a convex constraint obtained by performing a vision-based segmentation technique.

Although effective in obtaining good performance and robustness, the above optimization-based approaches have two main limitations from the viewpoint of this paper. First, classifying the terrain by employing different criteria, such as edges analysis, kinematics, and collision, can create a non-convex foothold constraint. This can be solvable by employing a Mixed Integer Program, which, however, can be hard to run in real-time~\cite{mixed_integer}. In fact, standard solutions, as in~\cite{mpc_vision2}, only consider the nearest convex constraint. Second, usually, disturbances are not explicitly estimated. In both cases, the obtained footstep candidate can be suboptimal. %Third, to perform accurate disturbance rejection via footstep correction, an online estimator is needed in the architecture, finally increasing the complexity of the controller.

To bypass the above limitations, Reinforcement Learning (RL) techniques can be viable solutions for achieving such optimality, since nonlinear and non-convex problems can be tractable by employing these methods. In~\cite{DeepLoco}, a footstep policy was learned from scratch together with a whole-body controller in a hierarchical fashion. Along the same lines, the authors in~\cite{deep_gait} proposed a similar architecture for the case of quadrupedal locomotion,
meanwhile, in~\cite{rl_hiking}, \cite{rl_spiaggia} RL was proposed to modify leg frequencies, and to generate foot position or joint residuals, showing impressive real-world results. In~\cite{rl_pandala}, RL was employed to add robustness to a model-based controller by estimating external disturbances.
Finally, similar to our approach, in~\cite{rloc} RL was used to learn a footstep planning policy while a model-based controller was employed to track the generated references. In the same way, here we employ an MPC controller to devise the robot motion, concentrating only on the footstep planning task.    

However, all these works do not explicitly consider safety in the learned policy, which can, in some scenarios, still fail given the black-box nature of the approach. This paper is aimed to directly take this aspect into consideration. %, proposing a new RL architecture to increase the safety of the final policy.%Furthermore, the final behavior of the policy is shaped by an extensive fine-tuning procedure of its parameters, which, given the number of reward functions that are usually considered, such as minimun distance with respect terrain edges, collision and kinematic reachability, enormously complicate in the considered task.
%{\bf CLAUDIO: I like this paragraph, it shows the delta of our method w.r.t. to the state of the art. To make our contributions crystal clear, add a paragraph that starts with "The main contributions of this work are:" followed by a list of 2-3 bullet points. Be specific and quantitative. e.g. write that our method reduced the violation of safety constraints by x\% compared to the state of the art, etc. The list of benefits below is nice, but it's not clear which or if any of the bullet points is a novel contribution.} 

In this paper, we build upon our visual foothold adaptation (VFA) technique~\cite{villarreal19ral} to devise a \textrm{safer} footstep planning policy for quadrupedal robots. The basic idea is to outsource the fulfillment of user-defined safety constraints by mixing the above model-based module with a specific policy architecture in order to guarantee with high probability their satisfaction. This is done by using the model-based priors given by the VFA, which perform terrain analysis and check for kinematic limits and shin collisions, both for defining the policy's input and modifying its output. The latter is done by employing a masking procedure over Proximal Policy Optimization (PPO)~\cite{PPO}, a  state-of-the-art model-free reinforcement learning algorithm, constraining the choice of the available actions to improve safety (Fig.~\ref{fig:scenarios}).
As a result, fewer reward terms (the ones that we relate to safety) are needed for shaping the behavior of the policy, which in return optimizes only dynamic conditions and terminal violations. 

%To summarize, the main benefits of the proposed approach are:
%\begin{itemize}
%    \item it applies to any quadrupedal and bipedal robot;
%    \item violation of user-defined safety constraints are greatly reduced, both during training and the final policy deployment;
%    \item additional safety constraints can be easily incorporated into the proposed approach; 
%    \item in general, fewer reward terms are required for shaping the behavior of the policy, enabling the possibility to achieve better performances with lower sample complexity.
%\end{itemize}

%As a comparison, first we show the benefit of adopting an RL policy instead of simpler heuristics for the choice of the foothold locations, and second we provide an extensive evaluation of the performance of our method with respect to standard RL approaches that minimize the same safety conditions in the reward.

\subsection{Contributions}
The main contributions of this work are:
\begin{itemize}
    \item the design of a footstep planning policy that incorporates a-priori safety information in its decision. To the best of the author’s knowledge, no prior work addresses this specific topic for footstep planning in the context of RL;
    %\item our approach can be applied to any quadrupedal and bipedal robot with minimal modifications;
    %\item with our approach, violations of user-defined safety constraints are greatly reduced, both during training and the final policy deployment; %furtermore, we propose a flexible architecture that can be easily modified to add additional constraints at demand; 
    %\item additional safety constraints can be easily incorporated into the proposed approach; 
    \item an extensive evaluation which shows that outsourcing safety translates into fewer reward terms for shaping the behavior of the policy, enabling the possibility of achieving better performances with lower sample complexity. Furthermore, we show that, with our approach, violations of the user-defined safety constraints are greatly reduced during the entire learning transient.
    %\item finally, we demonstrate that, in general, outsourcing safety translates into fewer reward terms for shaping the behavior of the policy, enabling the possibility of achieving better performances with lower sample complexity.
\end{itemize}

As a comparison, first we show the benefit of adopting an RL policy instead of simpler heuristics for the choice of the foothold locations, and second we provide an extensive evaluation of the performance of our method with respect to standard RL approaches that minimize the same safety conditions in the reward. 

%We conclude the paper by showing that, with our approach, violations of user-defined safety constraints are greatly reduced, both during training and the final policy deployment.

\subsection{Outline}
The paper is organized as follows. Section~\ref{sec:problem_formulation} introduces the footstep planning problem for quadrupedal robots, highlighting how vision-based and precise footstep placement can be used to perform locomotion in the presence of rough terrain. The proposed footstep planning policy is then presented in Sec.~\ref{sec:proposed_approach}, where we first discuss the learning procedure for the a-priori safety constraints, the policy architecture, and finally, the model-based controller adopted in this work. In Sec.~\ref{sec:results}, we report 
comparative simulation results, meanwhile, some general conclusions about the approach are drawn in Sec.~\ref{sec:conclusions}.

\begin{figure*}[!ht]
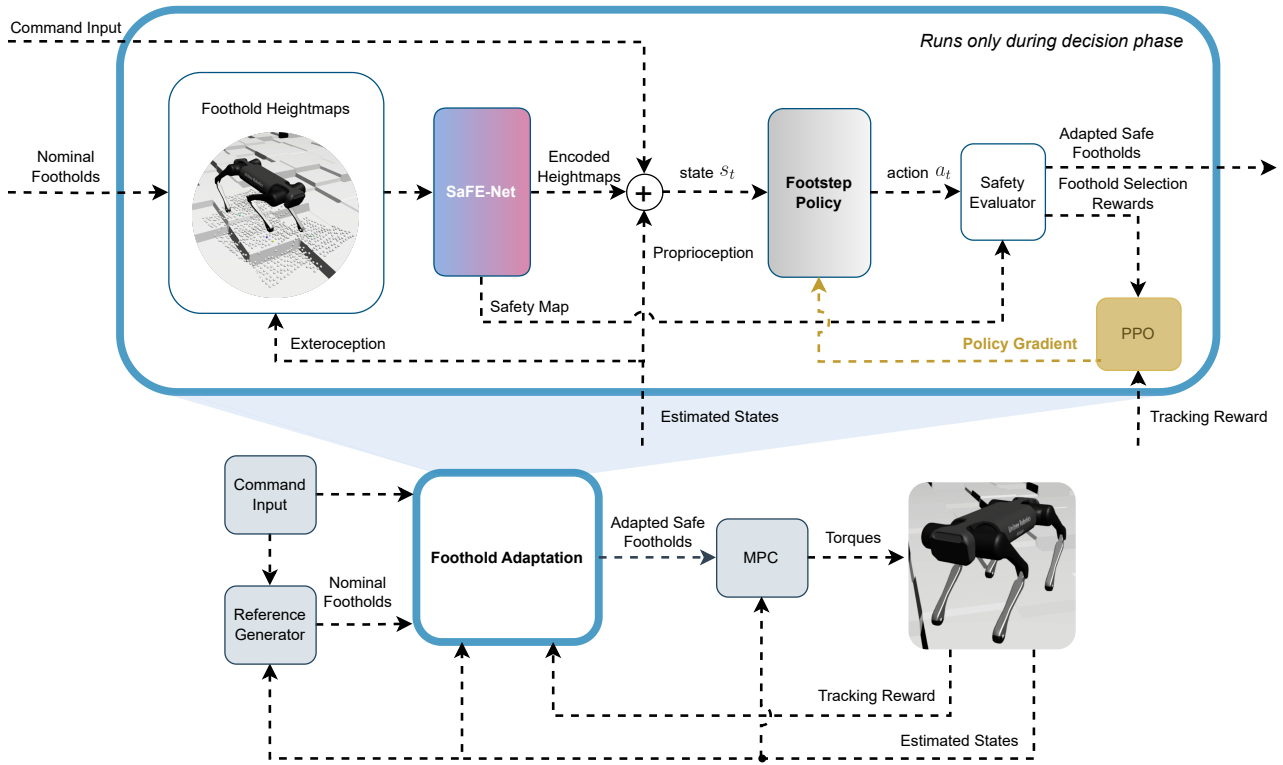

\BlockDiagram
\caption{Block diagram of the proposed approach. Starting from the left, the user commands a velocity input, which is then tracked by an MPC controller and an optimal footstep policy. The last, described in Sec.~\ref{sec:proposed_approach}, considers the safety information provided by SaFe-Net both in its input state and at the output level. }
\label{fig:block_diagram}
\end{figure*}

\section{Problem Formulation}
\label{sec:problem_formulation}
In this work, we want to learn a footstep planning policy that is able to choose optimal and safe foot landing locations in order to maximize the robot's performance. Commonly, in the case of blind locomotion and flat terrain, this is done by computing the next footstep $\pv_{\mathrm{foot}} \in \mathbb{R}^{3}$ as 
\begin{align}
    \pv_{\mathrm{foot}} = \boldsymbol{p}_{\mathrm{hip}}+\frac{1}{2}T_{\mathrm{stance}}(\boldsymbol{v}^{\mathrm{usr}}_{\mathrm{com}}+\boldsymbol{w}^{\mathrm{usr}}_{\mathrm{com}}\times { }^{\mathrm{b}}\boldsymbol{\bar{p}}_{\mathrm{hip}})
    \label{eq:raibert}
\end{align}
%by employing a modified version of the Raibert's heuristic~\cite{raibert}, 
where $\boldsymbol{\bar{p}}_{\mathrm{hip}} \in \mathbb{R}^{3}$ is the distance between the hip and center of the base,  $T_{\mathrm{stance}}$ is a scalar representing the stance duration, and $\boldsymbol{p}_{\mathrm{hip}} \in \mathbb{R}^{3}$ is the projection of the hip on the ground; finally, $\boldsymbol{v}^{\mathrm{usr}}_\mathrm{com}$, $\boldsymbol{w}^{\mathrm{usr}}_\mathrm{com} \in \mathbb{R}^{3}$ are respectively the linear and desired angular speed of the Center of Mass (CoM). All these quantities, if not explicitly specified, are expressed in the world frame $W$, meanwhile ${ }^{\mathrm{b}}\boldsymbol{\bar{p}}_{\mathrm{hip}}$ is expressed in the base frame $B$.

A drawback of the above equation is that it is only suitable for blind locomotion, and does not consider disturbances, limiting the robot's performance and resilience in a practical scenario.
For performing vision-based footstep planning, in this work, we build upon the VFA module which computes a set of heuristic criteria to evaluate the safety of each possible landing location in the vicinity of the proposed foothold $\pv_{\mathrm{foot}}$ in~\eqref{eq:raibert}. 

The VFA takes a tuple $\Tm$ as an input, evaluates it based on multiple criteria, and outputs a boolean matrix $\muv_{\text {safe }}$ representing which locations are considered safe. The input tuple for a generic leg in swing is defined as
$$
\Tm=\left(\Hm, z, \phi, {}^b\boldsymbol{v}_\mathrm{com}, {}^b\boldsymbol{w}_\mathrm{com}\right)
$$
where $\Hm \in \mathbb{R}^{h_x \times h_y}$ is the heightmap of dimensions $h_x$ and $h_y$ centered around $\pv_{\mathrm{foot}}$, and ${}^b\boldsymbol{v}_\mathrm{com}$, ${}^b\boldsymbol{w}_\mathrm{com} \in \mathbb{R}^{3}$ are respectively the linear and angular speed of the CoM in the base frame. Furthermore, $\phi \in [0,1]$ is a scalar representing the gait phase, which continually increases from 0 to 1 as the leg lifts off until the next touchdown, while $z \in \mathbb{R}$ is the hip height. Each cell of the heightmap $\Hm$ contains the terrain
height with respect to the hips of the robot.

%, and furthermore, $\Hm$ is oriented with respect to the horizontal frame of the robot. This frame coincides with the robot's base frame, and its $x y$-plane is perpendicular to the gravity vector. 

In this work, we only consider the following heuristic criteria: Terrain Roughness (TR), Leg Collision (LC), and Kinematic Feasibility (KF), which are detailed below:
\begin{enumerate}[label=\alph*)] 
\item Terrain Roughness (TR): this criterion checks edges in the heightmap that are unsafe for the robot to step on. For each candidate foothold $\pv_c$ in $\Hm$, we evaluate the height difference relative to its neighboring footholds, and we consider a threshold to decide whether $\pv_c$ is safe or not. 

\item Leg Collision (LC): this criterion selects footholds that do not result in a leg collision with the terrain at touchdown. To do so, we create a bounding region around the leg configuration corresponding to the candidate foothold $\pv_c$ and the future hip location. Then, if the bounding region collides with the terrain, we discard the candidate foothold.

\item Kinematic Feasibility (KF): this criterion selects kinematically feasible footholds, checking whether a candidate foothold $\pv_c$ will result in a trajectory that remains within the workspace of the leg during the entire gait cycle. For this, we check if the candidate foothold $\pv_c$ is within the workspace of the leg during touchdown and next lift-off. 

\end{enumerate}

 Each criterion $C$ outputs a boolean matrix $\muv_C$, and the final output $\muv_{\text {safe }}$ is computed by performing the element-wise logical AND $(\wedge)$ of all the criteria, such as 
 $$
\muv_{\mathrm{safe}}=\muv_{\mathrm{TR}} \wedge \muv_{\mathrm{LC}} \wedge \muv_{\mathrm{KF}}.
$$
The output $\muv_{\text {safe }} \in$ $\mathbb{R}^{h_x \times h_y}$ is a boolean matrix with the same size as the input heightmap $\Hm$ and indicates the safety of the candidate footholds.

In previous work~\cite{mpc_vision1}, we chose as the best foothold the one \textit{nearest} to the nominal $\pv_{\mathrm{foot}}$ flagged as safe in $\muv_{\text {safe }}$, but as detailed in the next section, this choice can result in a suboptimal robot behavior.

%being limited only on the above safety condition and the static criteria of eq.\textcolor{red}{[ref]}, which for example does not consider dynamic condition and possible disturbances. Hence, here we wish to further optimize the above choice by employing an RL policy, which taking into consideration the above safety criteria can only concentrate on optimizing a simpler velocity tracking cost in its reward function.

\begin{figure*}[!ht]
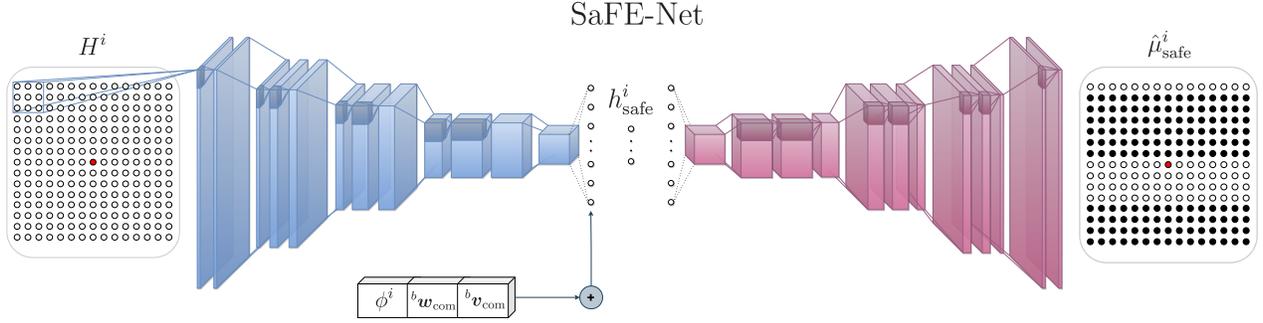

\SafeNet
\caption{Image representing the architecture of SaFE-Net (Sec.~\ref{sec:policy}). For a generic leg in swing $i$, on the left, we represent the heightmap $\Hm^i$, while on the right its prediction $\hat{\muv}^i_\text{safe}$. The center of the heightmap, $\pv^i_\text{foot}$ coming from~\eqref{eq:raibert}, is indicated with a red circle.  } 
\label{fig:safe_net}
\end{figure*}

\section{The Proposed Approach}
\label{sec:proposed_approach}

The presence of the robot's dynamics and disturbances affect considerably the footstep planning procedure. First, only applying the VFA module and choosing the nearest foothold location to $\pv_{\mathrm{foot}}$ can result in worse tracking performance of the user-commanded twist. For example, in the case of stair climbing, the robot could choose to step multiple times in the same area before proceeding. Second, even when the chosen foothold is optimal, the robot's motion can be destabilized by the presence of external disturbances, and another better step location might exist to aid a recovery maneuver.   

In this section, we describe our safe-footstep planning policy that is able to reduce the effect of the above conditions. At its core, there is a Neural Network (Sec.~\ref{sec:safe-net}) which classifies the heightmap patches $\Hm$ by learning the evaluation of the heuristic criteria described in Sec.~\ref{sec:problem_formulation} via supervised learning. The network employs an encoder-decoder structure to confine the classification information into a smaller subspace $\hv_\text{safe}$, which is then fed as input to an RL policy (Sec.~\ref{sec:policy}), which continuously proposes the best safe landing location by applying a masked variant of PPO. Finally, an MPC (Sec.~\ref{sec:control}) realizes the robot's motion.

A block diagram of the proposed approach is shown in Fig.~\ref{fig:block_diagram}.

\subsection{Visual Foothold Classification and Encoding} 
\label{sec:safe-net}
A common approach used to embed visual information inside an RL policy is by 
 employing a separate neural network with an encoder-decoder architecture~\cite{rloc}. This step is usually performed to compress the information into a smaller subspace to simplify the learning process, which otherwise will be more prone to local minima in the presence of redundant and unexplored state information.

In this work, we follow this same reasoning by condensing the VFA heuristics criteria detailed in Sec.~\ref{sec:problem_formulation} to give the policy a model-based prior on the footholds' safety. We perform regression via supervised learning by employing a denoising convolutional encoder-decoder style architecture, which we call as SaFE-Net (Safe Foothold Evaluation Network) to compute a fast and precise segmentation of the foothold heightmap $\Hm$. The encoder of SaFE-Net consists of a Convolutional Neural Network whose output is linearised and appended with some state information required to infer the safety of the foothold, followed by a linear layer, whose output is the encoded representation $\hv_\text{safe}$. The decoder consists of a linear layer, and its output is further reshaped and deconvolved to obtain a segmentation map $\hat{\muv}_\text{safe}$ where each cell corresponds to each foothold's safety. Fig.~\ref{fig:safe_net} provides a visual description of the architecture of SaFE-Net mentioned above.

%\sout{ For this purpose, starting from a dataset collected in RaiSim~\cite{raisim}, representing different heightmaps and their relative evaluation $\muv_\text{safe}$, we minimize a combination of the Binary Cross-Entropy~\cite{ml_book} (BCE) and Generalized Dice~\cite{dice} loss, such as}

To train SaFE-Net, we collect our heightmap dataset in RaiSim~\cite{raisim} over different rough terrain scenarios (see Fig.~\ref{fig:scenarios}). It consists of the foothold heightmap of the leg in swing $\Hm$, the subsequent safety evaluated using the heuristics $\muv_{\mathrm{safe}}$, the gait phase of the leg in swing $\phi$, the linear and angular velocities ${}^b\boldsymbol{v}_\mathrm{com}$ and ${}^b\boldsymbol{w}_\mathrm{com}$ of the center of mass. We minimise a combination of the Binary Cross-Entropy~\cite{ml_book} (BCE) and Generalized Dice~\cite{dice} losses, such as

\begin{align*}
    L_\text{BCE} = \frac{1}{N} &\sum_{i=1}^{N} ((\hat{\mu}_\text{i} \log\sigma(\mu_\text{i} + (1-\hat{\mu}_\text{i}) \log(1-\sigma(\mu_\text{i}))\\ 
    L_\text{dice} &= 1-(2 \cdot \frac{\sum_{i=1}^{N}(\mu_i \hat{\mu_i})} {\sum_{j=1}^{N} \mu_j + \sum_{d=1}^{N} \mu_d})
    \label{eq:loss}
\end{align*}
where $\hat{\mu}_i$ and $\mu_i$ are respectively the single element network's prediction and ground truth for the $i$-th element of the matrices $\hat{\muv}_\text{safe}$ and $\muv_\text{safe}$; $N$ is the number of elements in the flattened matrix $\Hm$, and $\sigma$ represent the sigmoid activation function. This combination of losses is commonly used in segmentation tasks to consider pixel-wise accuracy and global spatial coherence. The binary cross-entropy loss encourages accurate predictions at the pixel level, while the dice loss encourages spatially coherent segmentations by penalizing false negatives and false positives.

SaFE-Net provides two important components that form the basis of safety for training the subsequent footstep policy. The encoded representation of the heightmap, $\hv_\text{safe}$, which inherently embeds safety and is deployed as an additional state to the RL agent, and the segmented safety map $\hat{\muv}_\text{safe}$, which is used for action masking as described in the following section.

% In order to use this safety information in our footstep policy, the network architecture comprises of a small-width hidden layer $\hv_\text{safe}$, which we can extract during the policy training by simply querying the net and deploy as an additional state to the RL agent.

%twe take a compressed segmentation decision from a small-width hidden layer of the network, and furthermore we constrain the action space by masking the policy output by applying $\hat{\mu}_\text{safe}$. 

%The reader can refer to Fig.~\ref{fig:safe_net} for a visual description of the above architecture, to which we refer to as \textit{Safe-Net} in the rest of the paper.   

%Finally, to train the network, we collected the evaluation data of the VFA criteria in RaiSim~\cite{raisim}, a fast and parallelizable simulator that generates variegated rough terrains on demand. 

\subsection{Footstep Planning Policy}
\label{sec:policy}

The RL agent, acting as a centralized planner, determines the optimal footstep positions exclusively for the legs in swing, with the agent being queried only twice during this phase. These are during the lift-off of the foot and when it reaches the apex of the swing trajectory, which is generated by a separate model-based module. No more additional corrections are computed after the apex in order to avoid aggressive tracking maneuvers which can destabilize the robot's motion. 

The following describes the policy's state space, action space, the adopted safety masking procedure, reward functions, and training strategy.

\textit{1) State}:  The policy of our foothold adaptation strategy has states that represent both proprioceptive ($\xv_\text{prop}$) and exteroceptive ($\xv_\text{ext}$) information, such as 
\begin{align*}
    \xv_\text{prop} = ({}^b\vv^{\mathrm{usr}}_\text{com}, {}^b\wv^{\mathrm{usr}}_\text{com},  {}^b\vv_\text{com}, {}^b\wv_\text{com}&, ~\qv_\text{hist}, ~\Rm, ~\phiv^\text{legs}, ~\boldsymbol{a}_{\text{prev}}) \\
    \xv_\text{ext} = \hv_\text{safe}^\text{legs}
\end{align*}
%where ${}^b\vv_\text{usr} \in \mathbb{R}^{2}$ is the user-commanded $xy$-velocities and 
where $\qv_\text{hist} \in \mathbb{R}^{24\cdot t}$ is a sparse representation of $t$ past joint position and velocity values, which helps to reconstruct possible disturbance during motion; $\Rm \in \mathbb{R}^{3\times3}$ is a rotation matrix for retrieving the roll, pitch, and yaw of the robot; $\phiv^\text{legs} \in \mathbb{R}^{4}$ is a vector of phase variables; finally, $\boldsymbol{a}_{\text{prev}}$ is the previous action from the policy and $\hv_\text{safe}^\text{legs} \in \mathbb{R}^{32 \cdot 4}$ is obtained by stacking the encoded representations from SaFE-Net for the heightmaps of each leg. 

\textit{2) Actions}: The output of the policy is defined over a multidimensional continuous action space, where each possible action $\av^i$ is responsible for a precise $xy$-foothold coordinate for the $i$-th leg in the swing. In our case, the robot is constrained to perform a trotting gait. Hence only two legs can lift simultaneously, and the policy output $\av$ $\in \mathbb{R}^4$. The gait phases vector $\phiv$ helps the policy to understand which legs can be commanded at the current time instant.

\textit{3) Safety Masking}: After querying the agent, the continuous actions are converted to exact discrete foothold choices in the heightmaps $\Hm^i$ of the $i$-th legs currently in swing. One can normally assign a negative reward signal for each unsafe foothold choice to encourage safe foothold candidates. This approach, however, does not provide any safety guarantee for the policy to sample safe actions, even after the training procedure. Furthermore, unsafe foothold selections can happen rarely during training, hardening the optimization problem~\cite{sparse_RL}. To bypass the above limitations, we additionally employ an invalid action masking strategy to constrain the possible foothold's choice. Both during training and deployment, the output of SaFE-Net is employed to modify the action given by the policy if it is flagged as unsafe by choosing the \textit{nearest} foothold marked safe in $\hat{\muv}_\text{safe}$. 

\textit{4) Reward Functions}: Given the safety criteria considered in this work, we decided to employ a small set of reward functions to describe the desired behavior of the robot. These terms are related to the tracking performance of the user-commanded linear velocities ${}^b\vv_\text{usr}$ and a regularization behavioral component, defined respectively by the following functions:

\begin{align*}
    &\rv_\text{track} = \text{exp}\Bigg(-\frac{({}^b\vv^{\mathrm{usr}}_\text{com} - {}^b\vv_\text{com}))}{0.2\cdot(1 + |{}^b\vv^{\mathrm{usr}}_\text{com} - {}^b\vv_\text{com}|)}\Bigg)^2 \\
    &\;\;\;\;\;\;\;\;\;r_\text{reg} = \sum_{i=1}^{N_\text{swing}}|| \Hm^i(\av^i) -  \pv^i_\text{foot}||
\end{align*}
where, in $r_\text{reg}$, $\Hm^i(\cdot)$ is the foot position in the $i$-th heightmap proposed by the policy $\av^i$ converted to the corresponding discrete values, and $\pv^i_\text{foot}$ is the nominal foot position~\eqref{eq:raibert} at the center of the heightmap. Both these values refer to the $i$-th leg in swing. This reward term helps to stabilize the learning progress since we know that~\eqref{eq:raibert} is a good prior in the case of blind locomotion. Furthermore, we add some terminal sparse negative reward $r_\text{terminal}$ in the case of self-collision or other catastrophic behavior, such as hitting the ground with the trunk or exceeding some pitch and roll thresholds, and a negative constant reward in the case $\av^i$ is converted to unsafe footholds.

\textit{5) Training Procedure}: We trained our agent by applying external disturbances during random intervals to the CoM by sampling randomly in the range $[-35\text{N},35\text{N}]$, both in the longitudinal and lateral direction of the robot. We applied a curriculum strategy to increase the difficulty of the terrain and external disturbances during training. Furthermore, the terrain properties are randomized, and a new environment is sampled on every episode reset to encourage generalization. By employing PPO in the same simulation environment described in Sect.~\ref{sec:safe-net}, this training procedure is performed sequentially after a satisfactory precision of SaFE-Net is achieved. 

\begin{figure*}[!ht]
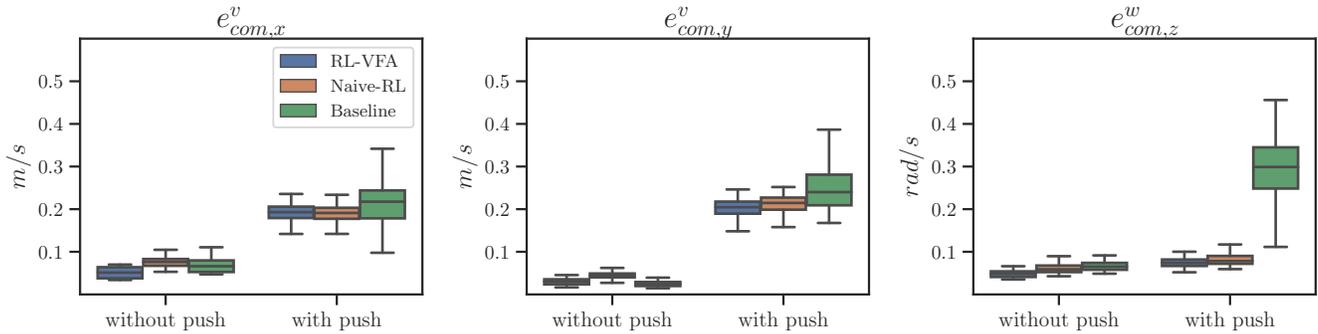

     %\centering
     %\begin{subfigure}[b]{0.8\textwidth}
         \centering
         \TrackingCost
         \caption{Median and variance of the linear ($x$-$y$) and angular (yaw) velocities tracking errors with and without the presence of external disturbances. These results were obtained by commanding the robot with different longitudinal velocities in the range of +/-~$0.3~m/s$.}
         \label{fig:tracking_x}
     %\end{subfigure}
     %\begin{subfigure}[b]{0.8\textwidth}
     %    \centering
     %    \TrackingCostY
     %    \caption{Tracking Cost with  Lateral velocity}
     %    \label{fig:tracking_y}
     %\end{subfigure}
    
     %   \caption{Tracking Costs?}
     %   \label{fig:tracking}
\vspace{-5pt}
\end{figure*}

\subsection{Model Predictive Control}
\label{sec:control}
To actuate the optimized foothold position chosen by the policy, we employ a well-studied formulation used for the real-time motion planning and control problem known as a Model Predictive Control (MPC). The formulation we utilized implements a linearized reduced order model to approximate the robot dynamics, called the Single Rigid Body Model (SRBM), which has shown its effectiveness, especially in the case of quadruped robots~\cite{SRBM} where the inertia of the legs can be usually neglected. Using this approximation, the system dynamics can be written as 
\begin{equation}
\label{eq:equation of motion}
	\begin{aligned}
		&\ddot{\boldsymbol{p}}_\text{com} = \frac{1}{m} \sum_{i=1}^{n_\text{leg}} \boldsymbol{f}_i + \boldsymbol{g}\quad \\
	&\dot{\boldsymbol{R}} = \boldsymbol{R}{}^b\hat{\boldsymbol{w}} \; \; \boldsymbol{R} \in SO(3)\\
  &{}^b\boldsymbol{I} {}^b\dot{\boldsymbol{w}} = \boldsymbol{R}^T(\sum_{i=1}^{n_\text{leg}}\hat{\boldsymbol{r}}_i \boldsymbol{f}_{i} ) - {}^b\hat{\boldsymbol{w}}  {}^b\boldsymbol{I} {}^b\boldsymbol{w}
	\end{aligned}	
\end{equation}
where $\ddot{\boldsymbol{p}}_\text{com}$ is the center of mass acceleration $\boldsymbol{f}_i$ is the Ground Reaction Force (GRF) acting on the $i$-th leg, $\dot{\boldsymbol{R}}$ is the derivative of rotation matrix between the world and the robot base, while $\boldsymbol{w}$ and $\dot{\boldsymbol{w}}$ are the angular velocity and its derivative. $\boldsymbol{r}_i$ is the vector connecting the base and foot $\boldsymbol{r}_i = \Hm^i(\av^i) - \boldsymbol{p}_\text{com}$, the operator $\widehat{(\cdot)}$ maps the vector to a screw-symmetric matrix. Finally, $\boldsymbol{I}$ and $m$ are the inertia matrix and mass of the robot body, and $g$ is the gravity vector. 
 Eq.~\eqref{eq:equation of motion} is non-linear in the angular part. We use the variation-based linearization scheme presented in~\cite{representation_free}. We express the rotational error in the $SO(3)$, considering the variation to the operating point to be free from singularities in the representation. We then perform a first-order Taylor expansion of the matrix exponential to then vectorize the error expressed in $SO(3)$ as $\boldsymbol{\xi} \in$  $\mathbb{R}^3 $ such that $\hat{\boldsymbol{\xi}} = \delta\boldsymbol{R}$. The linearized dynamics is finally discretized using the forward Euler scheme. The system state is defined as $ \boldsymbol{x} = \left[ \boldsymbol{p}_\text{com},  \dot{\boldsymbol{p}}_\text{com}, \boldsymbol{\xi}, {}^b\boldsymbol{w}\right] \in \mathbb{R}^{12}$ and the control input as 
$\boldsymbol{u} = \left[\boldsymbol{f}_1, \boldsymbol{f}_2, \boldsymbol{f}_3, \boldsymbol{f}_4 \right] \in \mathbb{R}^{12}$. Given the simplified model, system state, and control input we can define the optimal control problem (OCP) as: 
\begin{equation}
    \label{eq:optimization formulation}
    \begin{aligned}
        \min\limits_{\boldsymbol{x},\boldsymbol{u}} \quad &\mathcal{L}_T(\boldsymbol{x}(N))+\sum_{k = 0}^{N-1}{\mathcal{L}(\boldsymbol{x}_k ,\boldsymbol{u}_k)}\\
         s.t. \quad & \boldsymbol{x}_{k+1} = \boldsymbol{A}_{k}\boldsymbol{x}_k + \boldsymbol{B}_{k}\boldsymbol{u}_k + \boldsymbol{c}_k\\
         &\boldsymbol{u}_k \; \in \; U_k\\
         &k = 0,1, ...,N-1\\
        &\boldsymbol{x}_{k=0} = \boldsymbol{x}_\text{op} \\
    \end{aligned}
\end{equation}
where $\mathcal{L}(\boldsymbol{x}(\cdot))$ is a convex quadratic cost over the user commanded velocities ${}^b\vv^\text{usr}_\text{com}$, ${}^b\wv^\text{usr}_\text{com}$ and body posture. $\boldsymbol{A}_{k}, \boldsymbol{B}_{k}, \boldsymbol{c}_k$ are the linearized dynamics, and $U_k$ is the set of feasible ground reaction forces constrained by the outer pyramid approximation of the friction cone to guarantee non-slipping conditions. $\boldsymbol{x}_\text{op}$ is the state variable at the operating point. To solve the OCP problem we used a specialized quadratic programming solver~\cite{qpswift}, that exploits the sparse structure of the problem. 
Finally, the GRFs obtained by solving the optimization problem~\eqref{eq:optimization formulation} are then converted into motor torques by applying
\begin{equation*}
    \tauv = -\Jm^\top(\qv)\uv
\end{equation*}
where $\qv$ is the vector containing the actual robot joints position, and $\Jm(\cdot)$ is the contact jacobian.

\section{Results}
\label{sec:results}
The proposed approach has been validated through simulations on Aliengo~\cite{aliengo}, an electric quadruped robot developed by Unitree. In the following, we show simulation results by comparing three different approaches, which are
\begin{enumerate}
  \item MPC-VFA (baseline): the optimal foothold is chosen as described at the end of Sect.~\ref{sec:problem_formulation};
  \item RL-VFA: where we employ the masking procedure explained in Sect.~\ref{sec:policy}, encoding the safety information both in the policy input and output space; 
  \item Naive-RL: where the safety information is only encoded in the reward function as in the work of RLOC \cite{rloc}.
\end{enumerate}
To minimize the considered safety violations in the naive RL approach (3), we added some additional reward terms. Specifically, we designed two negative sparse reward terms in the case of kinematic violations and shin collisions, and a distance-based reward term if the chosen foothold is placed near a terrain edge. For the last computation, we use the same TR heuristic explained in Sect.~\ref{sec:problem_formulation}. Furthermore, similar to the footstep policy in~\cite{rloc}, we fed to the naive RL approach (3) an encoded representation of the heightmap (without providing any safety information) using a convolutional autoencoder architecture which was trained using the same heightmap data collected for SaFE-Net. The two policies mentioned above use the same training strategy as explained in Section III-B and an extensive hyperparameter search has been applied using Optuna~\cite{optuna} to optimize their results.

\begin{figure}[!b]
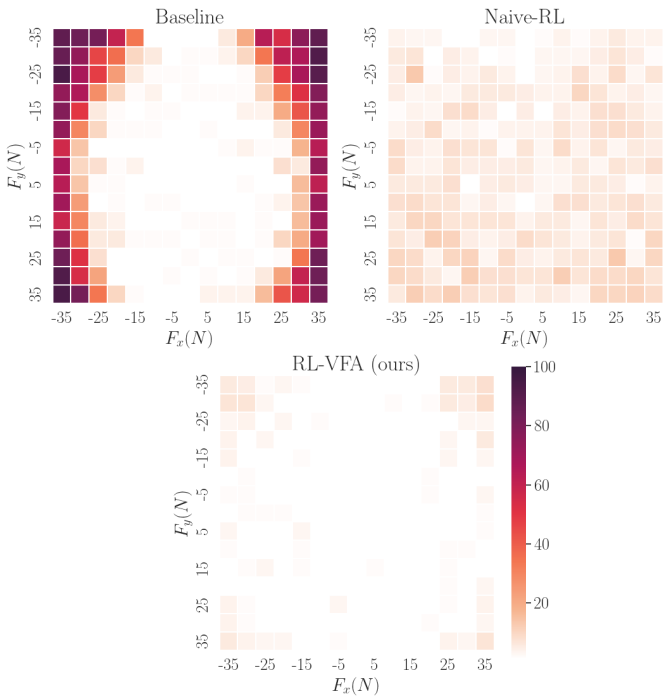

\vspace{-20pt}
\DisturbanceEval
\caption{Disturbance rejection analysis over external forces in the range $[-35\text{N}, 35\text{N}]$. Lighter squares depict higher robustness (zero episode termination in white). }
\label{fig:disturbance_eval}
%\vspace{-15pt}
\end{figure}

In the following, we want to highlight the benefit of employing a learned footstep planning policy within a model-based control framework and the performance increment that can be achieved by explicitly considering safety information. For this, we tested the robot over irregular and randomized terrains and averaged the obtained results over 100 episodes (each of them consisting of 250 footstep placements).

%As explained in Sect., a footstep planning policy that does not consider safety information can be more hard to optimize and still  (i.e. with a lower robustness of the robot) compared to a method that outsources these requirements; furthermore, we want to prove that learning-based method can outperform model-based alternatives in this footstep planning task.

%In all the results below we tested the robot over irregular and randomized terrains similar averaging the obtained result over 100 trials.

In Fig.~\ref{fig:tracking_x}, we compare the two learning methods against MPC-VFA, showing the linear and angular velocity tracking error $e^v_\text{com}$, $e^w_\text{com}$, achieved with and without the application of additional external disturbances. As shown in the figures, RL-VFA is able to achieve better results (in median and variance) compared to the other methods, reaching a mean reduction in the $x$ and yaw tracking error of $\approx$20\% compared to the baseline. A similar result can be observed in the presence of external disturbances due to the ability of the policy to recover faster after a push. Similarly, the simpler naive implementation of RL is able to achieve good tracking performance compared to the baseline. However, given the additional reward terms and the intrinsic difficulty in achieving the global optimum, this method achieved worse performance compared to our approach.

\begin{figure}[!t]
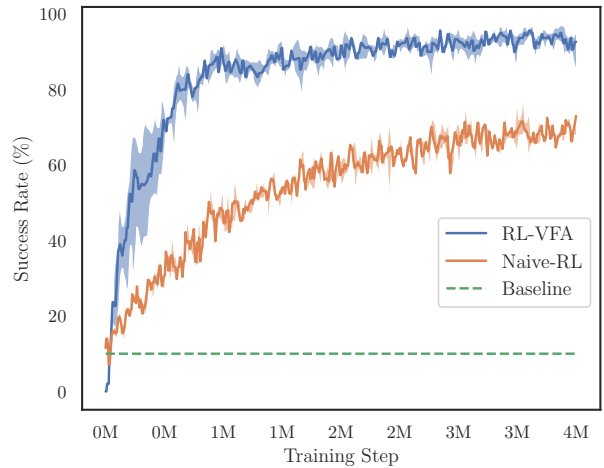

\vspace{-15pt}
\SuccessRate
\caption{Comparison of the episode success rate of Naive-RL and RL-VFA using three different hyperparameters during training on irregular terrains with fixed disturbances (+/-35N in the $x$-$y$ directions) applied at random intervals. To highlight the benefit of a learned footstep policy, we plot the mean results obtained by the baseline (MPC-VFA).}
\label{fig:success_rate}
\vspace{-10pt}
\end{figure}

This difficulty is clearly visible in the next result. To show that the proposed method can enable robust locomotion in the presence of severe disturbances, we analyzed the external forces at the CoM level withstood by all three methods (Fig.~\ref{fig:disturbance_eval}) applying them for a duration ranging from 1 to 3 seconds. The previous analysis can be applied here as well. Similar results are depicted in Fig.~\ref{fig:success_rate} where we plot the success rate (i.e. percentage of episodes that do not terminate with a body collision) achieved by all three methods during training. Here we superimpose the result coming from \textit{three of the best} hyperparameters found during our extensive hyperparameter search. The outsourcing of the safety information enables both better sample efficiency and a more robust locomotion. The last can be observed in the accompanying video, where shin collisions, kinematic violations, and slippage over terrain edges clearly require additional effort from the locomotion controller. 

Lastly, we compare in Table~\ref{table:violations} how the defined safety conditions hold in our approach and in the naive RL policy case. For this, we test both agents at the beginning of their training (10000 steps) and at their last episode (4000000 steps), and we calculate the percentage of safety violations over 100 episodes. Our approach consistently outperforms Naive-RL, especially in the case of terrain violations, where footholds are placed near a terrain edge $\approx$100 times more often compared to our approach. The same situation emerges in the case of shin collisions, where we achieved a reduction rate of $\approx$10x. Still, some violations remain even in our case given the inevitable classification error of SaFE-Net. These results can be explained by analyzing the number of violations of RL during the policy evaluation at the beginning of training: shin collision and kinematic violations are represented by sparse events, over which optimization is inevitably complicated. Given the sparse nature of these criteria, simply penalizing such actions that cause violations as in~\cite{rloc} is not necessarily sufficient to learn to avoid them. Furthermore, we want to highlight how there is no significant difference in the number of safety violations between the training episodes in our approach, since, given our masking procedure, their enforcement is not hinged on any particular learning curve. 

The reader can refer to the accompanying website\footnote{videos at: \href{https://sites.google.com/view/safe-steps-rl}{\textbf{https://sites.google.com/view/safe-steps-rl}}} for the videos of the above results. Furthermore, in the videos, we show another simulation on Gazebo~\cite{gazebo} and experiments on the real robot to prove that our method does not suffer from any particular domain adaptation problem. For this, we trained our policy by randomizing the robot's mass and inertia, and we added noise both to the MPC and to the policy states. Even in this case, we obtain similar comparative results to the ones described in this section.
\section{Conclusions}
\label{sec:conclusions}

In this paper, we have proposed a method for imposing safety 
constraints to a learned footstep planning policy via external model-based priors. The method hinges upon our visual foothold adaption technique (VFA) which classifies unsafe footstep locations by analyzing their kinematic feasibility, shin collision, and vicinity to terrain edges, information that is then used by the policy to plan only safe footholds. 

The proposed approach was statistically validated in simulation and tested on a real quadrupedal robot. In particular, numerical simulations of our method show that a low number of violations of the imposed safety conditions happen during training, resulting in an inherently safer footstep planning policy. Furthermore, we show how this approach is able to attain better final performances compared to standard RL approaches, thanks to a reduced number of reward functions needed to shape the behavior of the policy.

An interesting feature of the presented work is that additional constraints can be easily incorporated into the pipeline at demand just by modifying the training data for the encoder-decoder network. Given this feature, future work will consider the problem of coupling the proposed approach with dynamic constraints during the entire learning transient, for example, by explicitly taking into consideration criteria based on the Zero Moment Point~\cite{kajita} or the Instantaneous Capture Point~\cite{icp}, in order to increase the final robot's performance.

\begin{table}[ht]
\caption{Percentage of safety violations over 100 episodes}
\centering
\begin{tabular}{l|c|c|c} 
\hline 
\hline 
\textbf{Policy} & Terrain (TR) & Leg (LC) & Kinematic (KF) \\
\hline 
\rule{-2.75pt}{10pt}
RL-VFA (@10K steps) & 0.9  & 0.7   & 0.8 \\ 
RL-VFA (@4M steps)    & $\boldsymbol{0.6}$  & $\boldsymbol{0.3}$   & $ \boldsymbol{0.6}$    \\
Naive-RL (@10K steps)    & 48.4  & 5.8   & 1.77    \\
Naive-RL (@4M steps)    & 36.5  & 4.5   & 1.5    \\[0.5ex] 

\hline 
\hline 
\end{tabular}
\label{table:violations} 
\end{table}

\section{Acknowledgment}
This research is supported by and in collaboration with the Italian National Institute for Insurance against Accidents at Work (INAIL), under the project “Sistemi Cibernetici Collaborativi - Robot Teleoperativo 2”.
%\balance %could be skipped ADL

\bibliographystyle{IEEEtran}
\bibliography{bibliography}

\begin{thebibliography}{10}
\providecommand{\url}[1]{#1}
\csname url@rmstyle\endcsname
\providecommand{\newblock}{\relax}
\providecommand{\bibinfo}[2]{#2}
\providecommand\BIBentrySTDinterwordspacing{\spaceskip=0pt\relax}
\providecommand\BIBentryALTinterwordstretchfactor{4}
\providecommand\BIBentryALTinterwordspacing{\spaceskip=\fontdimen2\font plus
\BIBentryALTinterwordstretchfactor\fontdimen3\font minus
  \fontdimen4\font\relax}
\providecommand\BIBforeignlanguage[2]{{%
\expandafter\ifx\csname l@#1\endcsname\relax
\typeout{** WARNING: IEEEtran.bst: No hyphenation pattern has been}%
\typeout{** loaded for the language `#1'. Using the pattern for}%
\typeout{** the default language instead.}%
\else
\language=\csname l@#1\endcsname
\fi
#2}}

\bibitem{aliengo}
Unitree\phantom{-}Robotics, ``Aliengo,''
  \emph{https://www.unitree.com/aliengo}.

\bibitem{vision1}
J.~R. Rebula, P.~D. Neuhaus, B.~V. Bonnlander, M.~J. Johnson, and J.~E. Pratt,
  ``A controller for the littledog quadruped walking on rough terrain,'' in
  \emph{Proceedings 2007 IEEE International Conference on Robotics and
  Automation}, 2007, pp. 1467--1473.

\bibitem{vision2}
J.~Z. Kolter, M.~P. Rodgers, and A.~Y. Ng, ``A control architecture for
  quadruped locomotion over rough terrain,'' in \emph{2008 IEEE International
  Conference on Robotics and Automation}, 2008, pp. 811--818.

\bibitem{villarreal19ral}
O.~Villarreal, V.~Barasuol, M.~Camurri, L.~Franceschi, M.~Focchi, M.~Pontil,
  D.~G. Caldwell, and C.~Semini, ``Fast and continuous foothold adaptation for
  dynamic locomotion through cnns,'' \emph{IEEE Robotics and Automation
  Letters}, pp. 1--1, 2019.

\bibitem{book_mpc}
F.~Borrelli, A.~Bemporad, and M.~Morari, \emph{Predictive Control for Linear
  and Hybrid Systems}.\hskip 1em plus 0.5em minus 0.4em\relax Cambridge
  University Press, 2017.

\bibitem{mpc_vision1}
O.~Villarreal, V.~Barasuol, P.~M. Wensing, D.~G. Caldwell, and C.~Semini,
  ``Mpc-based controller with terrain insight for dynamic legged locomotion,''
  in \emph{2020 IEEE International Conference on Robotics and Automation
  (ICRA)}, 2020, pp. 2436--2442.

\bibitem{mpc_vision2}
R.~Grandia, F.~Jenelten, S.~Yang, F.~Farshidian, and M.~Hutter, ``Perceptive
  locomotion through nonlinear model-predictive control,'' \emph{IEEE
  Transactions on Robotics}, pp. 1--20, 2023.

\bibitem{mpc_safenet}
S.~Omar, L.~Amatucci, G.~Turrisi, V.~Barasuol, and C.~Semini, ``Fast convex
  visual foothold adaptation for quadrupedal locomotion,'' in \emph{4th Italian
  Conference in Robotics and Intelligent Machines (I-RIM)}, 2022, pp. 143--146.

\bibitem{mpc_vision3}
R.~Grandia, A.~J. Taylor, A.~D. Ames, and M.~Hutter, ``Multi-layered safety for
  legged robots via control barrier functions and model predictive control,''
  in \emph{2021 IEEE International Conference on Robotics and Automation
  (ICRA)}, 2021, pp. 8352--8358.

\bibitem{bratta}
N.~Rathod, A.~Bratta, M.~Focchi, M.~Zanon, O.~Villarreal, C.~Semini, and
  A.~Bemporad, ``Model predictive control with environment adaptation for
  legged locomotion,'' \emph{IEEE Access}, vol.~9, pp. 145\,710--145\,727,
  2021.

\bibitem{mixed_integer}
F.~Risbourg, T.~Corbères, P.-A. Léziart, T.~Flayols, N.~Mansard, and
  S.~Tonneau, ``Real-time footstep planning and control of the solo quadruped
  robot in 3d environments,'' in \emph{2022 IEEE/RSJ International Conference
  on Intelligent Robots and Systems (IROS)}, 2022, pp. 12\,950--12\,956.

\bibitem{DeepLoco}
X.~B. Peng, G.~Berseth, K.~Yin, and M.~van~de Panne, ``Deeploco: dynamic
  locomotion skills using hierarchical deep reinforcement learning,'' \emph{ACM
  Transaction on Graphics}, vol.~36, pp. 1--13, 2017.

\bibitem{deep_gait}
V.~Tsounis, M.~Alge, J.~Lee, F.~Farshidian, and M.~Hutter, ``Deepgait: Planning
  and control of quadrupedal gaits using deep reinforcement learning,''
  \emph{IEEE Robotics and Automation Letters}, vol.~5, no.~2, pp. 3699--3706,
  2020.

\bibitem{rl_hiking}
T.~Miki, J.~Lee, J.~Hwangbo, L.~Wellhausen, V.~Koltun, and M.~Hutter,
  ``Learning robust perceptive locomotion for quadrupedal robots in the wild,''
  \emph{Science Robotics}, vol.~7, 2022.

\bibitem{rl_spiaggia}
S.~Choi, G.~Ji, J.~Park, H.~Kim, J.~Mun, J.~H. Lee, and J.~Hwangbo, ``Learning
  quadrupedal locomotion on deformable terrain,'' \emph{Science Robotics},
  vol.~8, no.~74, p. 2256, 2023.

\bibitem{rl_pandala}
A.~Pandala, R.~T. Fawcett, U.~Rosolia, A.~D. Ames, and K.~A. Hamed, ``Robust
  predictive control for quadrupedal locomotion: Learning to close the gap
  between reduced- and full-order models,'' \emph{IEEE Robotics and Automation
  Letters}, vol.~7, no.~3, pp. 6622--6629, 2022.

\bibitem{rloc}
S.~Gangapurwala, M.~Geisert, R.~Orsolino, M.~Fallon, and I.~Havoutis, ``Rloc:
  Terrain-aware legged locomotion using reinforcement learning and optimal
  control,'' \emph{IEEE Transactions on Robotics}, vol.~38, no.~5, pp.
  2908--2927, 2022.

\bibitem{PPO}
J.~Schulman, F.~Wolski, P.~Dhariwal, A.~Radford, and O.~Klimov, ``Proximal
  policy optimization algorithms,'' \emph{ArXiv}, vol. abs/1707.06347, 2017.

\bibitem{raisim}
J.~Hwangbo, J.~Lee, and M.~Hutter, ``Per-contact iteration method for solving
  contact dynamics,'' \emph{IEEE Robotics and Automation Letters}, vol.~3,
  no.~2, pp. 895--902, 2018.

\bibitem{ml_book}
I.~J. Goodfellow, Y.~Bengio, and A.~Courville, \emph{Deep Learning}.\hskip 1em
  plus 0.5em minus 0.4em\relax MIT Press, 2016.

\bibitem{dice}
C.~H. Sudre, W.~Li, T.~K.~M. Vercauteren, S.~Ourselin, and M.~J. Cardoso,
  ``Generalised dice overlap as a deep learning loss function for highly
  unbalanced segmentations,'' \emph{Deep learning in medical image analysis and
  multimodal learning for clinical decision support : Third International
  Workshop}, vol. 2017, pp. 240--248, 2017.

\bibitem{sparse_RL}
D.~Rengarajan, G.~Vaidya, A.~Sarvesh, D.~Kalathil, and S.~Shakkottai,
  ``Reinforcement learning with sparse rewards using guidance from offline
  demonstration,'' in \emph{International Conference on Learning
  Representations}, 2022.

\bibitem{SRBM}
J.~Di~Carlo, P.~M. Wensing, B.~Katz, G.~Bledt, and S.~Kim, ``Dynamic locomotion
  in the mit cheetah 3 through convex model-predictive control,'' in \emph{2018
  IEEE/RSJ International Conference on Intelligent Robots and Systems (IROS)},
  2018, pp. 1--9.

\bibitem{representation_free}
Y.~Ding, A.~Pandala, C.~Li, Y.-H. Shin, and H.-W. Park, ``Representation-free
  model predictive control for dynamic motions in quadrupeds,'' \emph{IEEE
  Transactions on Robotics}, vol.~37, no.~4, pp. 1154--1171, 2021.

\bibitem{qpswift}
A.~G. Pandala, Y.~Ding, and H.-W. Park, ``qpswift: A real-time sparse quadratic
  program solver for robotic applications,'' \emph{IEEE Robotics and Automation
  Letters}, vol.~4, no.~4, pp. 3355--3362, 2019.

\bibitem{optuna}
T.~Akiba, S.~Sano, T.~Yanase, T.~Ohta, and M.~Koyama, ``Optuna: A
  next-generation hyperparameter optimization framework,'' \emph{Proceedings of
  the 25th ACM SIGKDD International Conference on Knowledge Discovery \& Data
  Mining}, 2019.

\bibitem{gazebo}
N.~Koenig and A.~Howard, ``Design and use paradigms for gazebo, an open-source
  multi-robot simulator,'' in \emph{2004 IEEE/RSJ International Conference on
  Intelligent Robots and Systems (IROS)}, vol.~3, 2004, pp. 2149--2154 vol.3.

\bibitem{kajita}
S.~Kajita, H.~Hirukawa, K.~Harada, and K.~Yokoi, ``Introduction to humanoid
  robotics,'' in \emph{Springer Tracts in Advanced Robotics}, 2014.

\bibitem{icp}
J.~Pratt, J.~Carff, S.~Drakunov, and A.~Goswami, ``Capture point: A step toward
  humanoid push recovery,'' in \emph{2006 6th IEEE-RAS International Conference
  on Humanoid Robots}, 2006, pp. 200--207.

\end{thebibliography}

\end{document}